\documentclass[journal, 12pt, onecolumn]{IEEEtran}

\usepackage{adjustbox}
\usepackage{amssymb}
\usepackage{amsthm}
\usepackage[cmex10]{amsmath}
\usepackage{amsfonts}
\usepackage{algorithmic}
\usepackage[linesnumbered, ruled, vlined, onelanguage, ]{algorithm2e}
\usepackage{array}
\usepackage{authblk}
\usepackage[english]{babel}
\usepackage{blindtext}
\usepackage{booktabs}
\usepackage{cite}
\usepackage{colortbl}
\let\labelindent\relax
\usepackage{enumitem}  
\usepackage{eqparbox}
\usepackage[T1]{fontenc}
\usepackage[hang,flushmargin]{footmisc}
\usepackage{graphicx}
\usepackage{import}
\usepackage{latexsym}
\usepackage{lipsum}
\usepackage{listings}
\usepackage{hyperref}
\hypersetup{
  colorlinks   = true, 
  urlcolor     = black, 
  linkcolor    = black, 
  citecolor   = black 
}
\usepackage{mdwmath}
\usepackage{mdwtab}
\usepackage{multicol}
\usepackage{multirow}
\usepackage{pifont} 
\usepackage[tight,footnotesize]{subfigure}
\usepackage{stfloats}
\usepackage{supertabular}
\usepackage[dvipsnames,table]{xcolor}
\usepackage{xparse}

\usepackage{silence}
\usepackage{caption}

\definecolor{Gray}{gray}{0.9}
\definecolor{LightCyan}{rgb}{0.88,1,1}
\usepackage[first=0,last=9]{lcg}

\newcommand{\vsp}{\vspace{1em}}
\renewcommand{\indent}{\hspace{0.7cm}}
\rowcolors{2}{gray!25}{white}
\newcommand*\rot[1]{\rotatebox{90}{#1}}

\begin{document}

\title{\textcolor{white}{.}\\ Individual Explanations in Machine Learning Models: A Case Study on Poverty Estimation }
\author{\IEEEauthorblockN{
\hspace{0.5cm} Alfredo Carrillo \hspace{1.8cm}
Luis F. Cant\'u  \hspace{1.8cm}  
Luis Tejerina   \hspace{0.9cm}
Alejandro Noriega}

\IEEEauthorblockA{
alfcar9@prosperia.ai \hspace{1cm}
cantu@prosperia.ai \hspace{1cm}
luist@iadb.org   \hspace{1cm}
noriega@mit.edu}\\
}

\providecommand{\keywords}[1]{\textbf{\textit{Index terms---}} #1}

\maketitle
\thispagestyle{empty}

\begin{abstract}
Machine learning methods are being increasingly applied in sensitive societal contexts, where decisions impact human lives. Hence it has become necessary to build capabilities for providing easily-interpretable explanations of models' predictions. Recently in academic literature, a vast number of explanations methods have been proposed. Unfortunately, to our knowledge, little has been documented about the challenges machine learning practitioners most often face when applying them in real-world scenarios. For example, a typical procedure such as feature engineering can make some methodologies no longer applicable. The present case study has two main objectives. First, to expose these challenges and how they affect the use of relevant and novel explanations methods. And second, to present a set of strategies that mitigate such challenges, as faced when implementing explanation methods in a relevant application domain---poverty estimation and its use for prioritizing access to social policies.
\end{abstract}

\small
\hspace{0.75cm}
\normalsize

\tableofcontents

\pagebreak
\section{Introduction}

\vsp
\subsection{Relevance of Model Explanations in Real-World Contexts}
\label{relevance}

Complex estimation and decision-making tasks have traditionally been analyzed and judged by human experts. Hence, decisions have typically been able to be complemented with human-interpretable justifications, when needed, as experts can normally explain the line-of-thought that led to their own decision-making. However, in the past two decades, algorithmic decision-making has spread increasingly to many relevant societal contexts. Despite the notable enthusiasm for the potential benefit that this type of technology can bring, the underlying methods used are typically not inherently transparent, in the sense that they do not readily provide human-interpretable justifications for their decisions \cite{molnar2019interpretable}. Moreover, in recent years there is a trend where the most successful algorithms, particularly in complex tasks like machine vision and natural language processing, tend to rely on highly complex models, which has led to a further increase in tension between accuracy and interpretability \cite{Goodfellow-et-al-2016}.

\vsp
Relevant societal contexts where algorithmic decision systems have gained substantial traction include medical diagnosis and treatment  \cite{kleinberg2015prediction}, counter-terrorism \cite{pelzer2018policing}, criminal justice \cite{kleinberg2016inherent}, and risk assessments for credits and insurance \cite{wan2011prediction}. In such impactful contexts, there is a legitimate need for providing human-interpretable explanations along with the estimations and decisions made. Indeed, lack of interpretability has become a barrier to the adoption of machine learning-based systems in many institutions and companies. Hence the value of complementing ML models with human-interpretable accounts of the statistical rationals behind their estimations, in a way that human decision-makers can more easily understand machine estimations, and even integrate their statistical rationals with qualitative information and human expert judgements. To address this challenge and opportunity, in recent years the academic literature has proposed a large number of ML methods for interpretability and explainability\footnote{ Terms which we use interchangeably in this work}. 

\vsp
In this work we assess the applicability of state-of-the-art explainability methods in the face of common real-world challenges faced by machine learning (ML) practitioners, and illustrate the challenges and a set of mitigation strategies as applied to a relevant application domain: poverty estimation and its use for prioritizing access to social policies \cite{noriega2020algorithmic}.

\subsection{Poverty Estimation and Prioritizing Access to Social Policies}
\label{poverty}

Most social policies, like cash transfers, are targeted at the poor. However, in the developing world, reliable income data is typically not available and costly to procure because most people work in informal economic markets \cite{bonnet2019women}. Hence, targeting commonly relies on algorithms that estimate households’ poverty based on observable and less costly proxy data, such as education levels, demographics, and their assets and services \cite{center2009manual, fiszbein2009conditional, hanna2018universal, ibarraran2017conditional, bank2018state}. In most countries, these algorithms are trained to estimate poverty based on large, periodic, statistically representative household surveys, which collect both the proxy features and the income ground truth. In their daily operations, institutions cannot collect ground truth income data directly from potential beneficiaries, due to the cost of eliciting trustworthy income data at such a massive scale, as well as candidates’ strong incentives for under-reporting.

\vsp
In practice, estimation models are imperfect, leading to targeting errors. In Latin America and other regions, it is estimated that targeting systems incur more than 25\% exclusion errors and inclusion errors (undercoverage and leakage) \cite{ibarraran2017conditional, mcbride2018retooling}. However, the methods used during the past two decades for estimating poverty have relied on econometric approaches that are not optimized for out-of-sample prediction, and cannot leverage the many non-linear relationships that are typically found in high-dimensional data. Hence, it has been shown that substantial accuracy gains can come from the use of modern computational and statistical methods borrowed from the field of ML; for example, extending the coverage of the poor by nearly a million people in two countries, without increasing expenditure \cite{noriega2020algorithmic}. The present work focus on the task of building accurate household income estimation models, such as in the work of Noriega et al. from 2020 \cite{noriega2020algorithmic}, but additionally providing them with a layer of explainability.

\subsection{Interpretability of Poverty Estimation Models}

 As mentioned in Section \ref{poverty}, even when ML models can be used to produce fair and accurate poverty estimations, it is highly desirable to provide human-interpretable accounts of the statistical rationals behind a model's estimations. For example, i) the head of the households may require to know the rationals which justify why they were not prioritized to a given social program, and/or, ii) social workers may need a way to integrate the statistical rationals behind a model's estimation, together with additional qualitative information of the household, for reaching a final improved judgement about that case.    

\vsp
In our work, we have built a series of statistical models based on methods of increasing complexity. We began with \textit{so-called} intrinsic interpretable models---such as linear regressions and single tree regressors---and then proceeded to models with higher complexity---such as random forests and gradient boosting regressors \cite{noriega2020algorithmic}. It is commonly noted in literature that in tasks with high-dimensional input, as we move towards higher complexity, interpretability is lost, while accuracy is gained \cite{Goodfellow-et-al-2016, nielsen2016tree}. Others sustain that not necessarily \cite{shrikumar2016not}. In our particular case, the most accurate model (a gradient boosting regressor) was not intrinsically interpretable, and the accuracy drop between this model and all interpretable models was substantial. Thus, we kept the not-intrinsically interpretable model and decided to explain it via Model-Agnostic Explanation Methods, which add a layer of interpretability between the model and potential human audits \cite{carrillo2021individual}. 

\vsp
Section \ref{real} relates a relevant set of available methodologies with a set of challenges that appear when implementing such methodologies in real-world scenarios, and which are particularly present in our application domain. Section \ref{proposed} proposes an ad-hoc methodology that effectively addresses such challenges by introducing a set of mitigation strategies appropriate to our case study. In Section \ref{platform} we describe briefly an interactive platform we developed to explain, and visualize individual explanations. Section \ref{conclusion} concludes this case study.

\section{Real-World Challenges vs. Available Methodologies}
\label{real}

\subsection{Scope of Methodologies}

Recently, there has been a notorious increase in the number of methodologies proposed for ML interpretability. They encompass numerous purposes, and each contains several method alternatives. In this topic, there are standard taxonomies used in literature: (i) Global vs. Local, (ii) Model-Agnostic vs. Model-Specific, or (iii) Post-Hoc vs. Intrinsic Interpretable Models. The scope of the methodologies reviewed here can be characterized as the following class intersection:

\begin{itemize}
    \vsp
    \item \textit{Local (Individual Explanations)}. Any model that provides individual explanations is local. In contrast with Global-Explanations, these methods focus on a small neighborhood rather than the whole feature space. This neighborhood has as centroid an observation of interest which we wish to explain. Perhaps the model is complex to understand fully, but predictions can be explained by a simple set of rules within a neighborhood.
    
    \vsp
    \item \textit{Post-Hoc.} In contrast with the Intrinsic Interpretable Models, these methods add an explainability layer between high complexity degree ML models and humans. The addition of this layer is regarded as an independent step of the training and testing of the model.
    
    \vsp
    \item \textit{Model-Agnostic}. These methods apply to any ML model, as they separate models from the explanations \cite{molnar2019interpretable}. Since they cannot access the internal parameters, they analyze it via a sample of inputs and corresponding outputs to assess its internal mechanisms. Because every Model-Agnostic Method is Post-hoc, it is simpler to define our scope as the intersection of Local and Model-Agnostic classes only, but we did not so for completeness.
    
\end{itemize}

\vsp
We selected such a framework because (i) we needed to provide individual explanations for the member's household; (ii) the interpretability is post hoc; (iii) and by using model-agnostic methods, we could apply them to any ML model, the methodology could be repeated in other countries.

\subsection{Real-World Challenges}
\label{real_sub}
In theory, the implementation of these methods should be straightforward. But empirically, most of them incur different kinds of problems that are complicated to fix. For example, with specific conditions of the data set---which are described further in this section---these methods would likely provide deficient explanations. Other methods provide interpretability in a way that it is not desirable for this context. Indeed, solving---or mitigating---these problems are \textit{challenges} for \textit{real-world} scenarios. 

\vsp
It is convenient to introduce the different real-world challenges according to how they affect two major types of approaches existing within model-agnostic methods: Perturbation Approach and Contrastive Approach. Here we describe these two approaches and expose how the severity of a challenge is closely related to the method's approach.

\vsp
\textbf{Perturbation Approach.} A widely popular approach is based on attributing the prediction output as individual effects of each of its input features, and it aims to find the most dominant ones \cite{lundberg2017unified, ribeiro2016model}. For estimating each feature's effect, a perturbation approach is commonly followed. A sample of artificial instances is generated via observed instances in the data set by distorting---or perturbing---slightly a few numbers of features while leaving the remain unchanged. The specific way to generate the sample varies across methods. In this particular context of individual explanations, the perturbations are centered on the instance of interest.

\vsp
The most common and salient challenges of methods that follow a Perturbation Approach are:
\begin{enumerate}
    
    \vsp
    \item \textbf{Unlikely or Impossible Observations.} If features are highly correlated, perturbations must be carefully crafted. Otherwise, unlikely or impossible observations could be generated. The model would likely err the prediction for these instances because it has never seen such combinations of feature values. Since the predictions are used for explainability purposes, the interpretability method would probably be inaccurate. This challenge is discussed in detail in Section \ref{unlikely}.
    
    \vsp
    \item \textbf{Contrastive Explanations.} A desirable property is to make contrastive explanations as humans understand a certain decision better when contrasted against others. The challenge consists of giving a contrastive interpretation of these methods and is examined further in Section \ref{contrastive_sub}.
    
    \vsp
    \item \textbf{Non-Interpretable Model Input.} The preprocessed representation of instances in ML models is non-interpretable for humans; One-Hot-Encoding is a typical example of this representation, in which a classification feature of $n$ bins is transformed into $n$ binary columns---or $n-1$ because of redundant information. Providing explanations in terms of these features would be useless for humans. This challenge is discussed thoroughly in Section \ref{non}.  
    
\end{enumerate}

\vsp
\textbf{Contrastive Approach.} It explains by contrasting a focal instance against a close by reference group. Some methods use the same label prediction for the reference group and focal instance, while others use a different one. In the former fashion, the explanation consists on demonstrating that the focal instance belongs to a cluster in which similar predictions are made. Whereas for the latter, the explanations are counterfactual. The seek for the smallest changes of the focal instance such that the prediction label changes. These are simple explanations and useful to offer a recourse for users who obtained unfavorable decisions \cite{garg2019counterfactual}.

\vsp
Because our case-study method followed a perturbation approach, each of these challenges is further analyzed in a later section, with its correspondent mitigation strategies. Whereas, the Contrastive Approach challenges are listed below but not further reviewed.

\vsp
The most common and salient challenges of methods that follow a Contrastive Approach are:
\begin{enumerate}
    \vsp
    \item \textbf{Distance Definition.} A distance measure is used to test the similarity between instances. In real-world scenarios, where there are different types of data: numerical---with different scaling---categorical and booleans, it is cumbersome to define a distance measure with such a diversity of data types.
    
    \vsp
    \item \textbf{Curse of Dimensionality.} The curse of dimensionality is a high dimensional space phenomenon where instances tend to be sparse---too far apart from each other. The nearest neighbors of a focal instance can be so distant that it is no longer reliable to regard them as ``similar observations''. 
   
    \vsp
    \item \textbf{Optimization Problem.} For the particular case of counterfactuals, the optimization problem for finding the closest data-point such that the prediction label changes is \textit{NP-Hard}, which means there is no known efficient algorithm that can solve it within a reasonable lapse of time. Also, the solution must provide an instance real-world observable, i.e. feasible. 
    
    \vsp
    \item \textbf{Unreliable Counterfactuals.} Again, in the case of counterfactuals, but specifically for classification problems, there is a latent risk of making counterfactual explanations based on counterfactual points that are not reliable. A discussion and mitigation of this challenge can be found at \cite{laugel2019dangers}.
   
\end{enumerate}

\begin{table}[ht!]
\caption{Available Model-Agnostic Methods vs Challenges}
\resizebox{0.70\columnwidth}{!}{
\label{methods}
\begin{tabular}{p{1.5cm}c|l|c|c|p{0.5cm}|p{0.5cm}|p{0.5cm}|c}
  \cmidrule{3-8}
    \ & & Method Name                       & Year & Article                           & \multicolumn{3}{c|}{Severity of Challenge}\\
    & &                                   &      &                                   & \ref{unlikely}   & \ref{contrastive_sub}   & \ref{non}   \\ \cmidrule{3-8}
    & \multirow{11}{*}{\rot{\textbf{Perturbation-Based}}}
      & PDP                              & 2001 & \cite{friedman2001greedy}         & \ H & \ H & \ H \\
    & & ICE                              & 2015 & \cite{goldstein2015peeking}       & \ H & \ H & \ H \\
    & & ALE                              & 2016 & \cite{apley2016visualizing}       & \ M & \ H & \ H \\
    & & Shapley Values (SHAP)            & 2017 & \cite{lundberg2017unified}        & \ M & \ H & \ H \\
    & & LOCO                             & 2018 & \cite{lei2018distribution}        & \ H & \ H & \ H \\
    & & Decomposition of prediction      & 2008 & \cite{robnik2008explaining}       & \ H & \ H & \ H \\
    & & Feature Importance               & 2018 & \cite{casalicchio2018visualizing} & \ M & \ H & \ H \\
    & & Sensitive Analysis               & 2013 & \cite{cortez2011opening}          & \ H & \ H & \ H \\
    & & LIME                             & 2016 & \cite{ribeiro2016model}           & \ H & \ H & \ H \\
    & & Explanations Vectors             & 2010 & \cite{baehrens2010explain}        & \ H & \ H & \ H \\
    & & Anchors                          & 2018 & \cite{ribeiro2018anchors}         & \ H & \ H & \ H
    \\\cmidrule{3-8}
    & \multirow{3}{*}{\rot{\scriptsize{\textbf{Contrast}}}}
    & Counterfactuals & 2017 & \cite{wachter2017counterfactual}                       & \ H & \ L & \ M \\
    & & Prototype and Criticism            & 2016 & \cite{kim2016examples}            & \ H & \ L & \ M \\
    & & Justified Counterfactual           & 2018 & \cite{laugel2019dangers}          & \ M & \ L & \ M \\ \cmidrule{3-8}
    & \multicolumn{8}{l}{\textit{\ \ \ref{unlikely}: Unlikely or Impossible Observations, \ref{contrastive_sub}: Contrastive}}\\
    & \multicolumn{8}{l}{\textit{\ \ Explanations, \ref{non}: Non-Interpretable Model Input}}\\
    & \multicolumn{8}{l}{\textit{\ \ H: High Severity, M: Moderate Severity, L: Low Severity}}\\
\end{tabular}
}
\end{table}

\pagebreak
In Table \ref{methods}, we considered the same list of the most relevant and novel Model-Agnostic Methods for individual explanations considered in \cite{carrillo2021individual}. We determined for each, whether the Severity of the Perturbation Approach Challenges is High (H), Moderate (M), or Low (L). The labels of the challenges, \ref{unlikely}, \ref{contrastive_sub}, and \ref{non}, correspond to the Section name where each challenge is further discussed. The upper part corresponds to Perturbation Approach methods, while the lower to Contrastive Approach methods. 

\vsp
The approach class is closely related to the degree of severity of the challenge. Almost all reviewed methods suffer highly from \ref{unlikely}; the exceptions are discussed in that section. Naturally, the Challenge \ref{contrastive_sub} has High Severity for Perturbation Methods and Low for Contrastive Methods. Similarly, the Challenge \ref{non} has High Severity for perturbation methods, whereas Moderate Severity for the contrastive methods because these methods have potential for transforming the Feature Space into an Interpretable Space.

\subsection{Unlikely or Impossible Observations}
\label{unlikely}
In real-world data sets, features tend to have a degree of correlation. If we overlook this fact, then it is likely that a collection of the artificial instances randomly generated can hardly ever happen in the real-world scenarios. Specifically, the model would receive combinations of feature values that has never seen before. Evidently, when the model predicts outcomes for these instances, they are prone to be biased or unconfident. If the features' effects are calculated via these values, then the method's confidence of interpretability is inherently low. Therefore, the perturbation stage in real-world databases must be carefully crafted.

\vsp
There are two types of correlations across features: 1) Underlying Distribution of Probability and 2) Feature Dependencies. Some of the mitigation strategies for this challenge can be applied to only a particular type. 

\begin{enumerate}
    \vsp
    \item \textbf{Underlying Distribution of Probability.} An unknown Distribution of probability underlies the correlation across features. In this type of correlation, it is unknown how or to what extent features are correlated. It is the most relevant challenge described in this work, as it is possibly the challenge that most increases the probability of providing unreliable and unintuitive explanations of the model.
    
    \vsp
    \item \textbf{Feature Dependencies.} They occur when features have a direct and known relation. Instead of an underlying Distribution of probability, the relation across features can be understood as a deterministic mathematical function. To generate an artificial instance, if one does not respect these acknowledged relations, it is possible to obtain an impossible or inconsistent observation. In this circumstance there is certainty that the model was not trained with instances that had such combinations of values. Moreover, these are instances outside of the Feature Space. There are two types of feature dependencies: 

\vsp
\begin{enumerate}[label=(\roman*)]
    \item \textbf{Nested Features.} In contexts such as the households one, they occur when the database contains nested features. For example, in data sets collected from surveys, each feature corresponds to a particular question. However, not every individual is asked the same questions. Surveys explicitly have question dependencies, where according to a specific answer, it is indicated if the next question should be responded or not. Or even more drastically, which sections should be answered and which others should be skipped. For example, in a socio-economic study, a minor of age would skip all the questions corresponding to labor, whereas the adults would likely skip other kinds of questions. 
    
    \vsp
    \item \textbf{Preprocessed data.} This kind of dependency arises in cases where preprocessed data---or Feature Engineering---is employed. Almost every real-world database requires a preprocess before an ML model can be applied. In this stage,  expert variables are crafted from the initial features, providing explicitly to the model patterns in the data that ad-hoc are important or useful for improving its accuracy. Other common preprocessing techniques are One-Hot Encoding, binning, and grouping operations. Here features are designed for the machine's processing. For example, features derived from binning or One-Hot-Encoding, have a strict acknowledged dependency. 
\end{enumerate}    
\end{enumerate}

\vsp
In these contexts, the moral is that it is required a mechanism to generate carefully a sample of instances in which each has a positive probability of being real-world observable---or feasible---while not necessarily being observed directly in the training data set.

\vsp
\textbf{Mitigation Strategies.} 

\begin{enumerate}
    \vsp
    \item \textbf{Use of the Observed Distribution:} During the generation of artificial instances, in each iteration is determined which features are perturbed and which values are used for replacement. A powerful idea is drawing values from the observed distribution, rather than generate them randomly. By doing so, each artificial instance is more likely to appear in the real-world. Several strategies follow this approach:
    
    \begin{enumerate}[label=(\roman*)]
        \vsp
        \item \textbf{Use of Conditional Distributions.}
        For preventing the generation of artificial instances that are unfeasible, ad-hoc conditionalities are defined across features. After perturbing a particular feature-value, the next values for replacement are drawn from a smaller sampling space, defined by the conditional distribution of observing the first replaced value. The conditionalities are generally implemented only in those features that have high correlation across numerous features, such as age or sex. This strategy was implemented in the Algorithm of Section \ref{add_conditional}.
    
        \vsp
        \item \textbf{Use of the Natural Distribution for Bivariate perturbation.} Commonly a univariate perturbing approach is followed because it helps to understand the feature-wise effects. But even if the values are drawn from the observed distribution, it still overlooks the correlation across features---particularly the highly correlated ones. This problem can be mitigated by increasing the number of features that are simultaneously perturbed. Unfortunately, this raises the method's computational cost. A middle point alternative is to use a bivariate approach. It is not as costly as modifying an arbitrary number of features, and it mitigates the problem of correlated features. This strategy implementation is discussed in detail in Section \ref{bi_sub}.
        
        \vsp
        \item \textbf{Use of Combinations of Simultaneously-Observed Values for Replacement.} This strategy is used in the article by Lipovetsky \cite{lipovetsky2001analysis} for Shapley Values estimation. During the artificial instance generation, in each iteration and assuming multiple values are replaced, instead of drawing them independently from the data set, a random instance is drawn. The corresponding feature values of this instance are used for replacement. By doing so, the artificial observations are likely more observable. The work of Feature Importances in \cite{casalicchio2018visualizing} proposed a permutation-based on SHAP, which also enables it to mitigate this challenge with a similar procedure. That is why both the Shapley Values and Feature Importances are Moderately affected by this challenge, as shown in Table \ref{methods}. 
        
    \end{enumerate}
    
    \vsp
    \item \textbf{Use of Differences to Eliminate Correlation.}
    Apley et al. \cite{apley2016visualizing} proposed a graphic method called ALE plots for a scenario with correlations across features. They use the differences in predictions, instead of using averages, to block the features' interactions. By doing so, they mitigate problems that appeared in methods such as PDPs \cite{friedman2001greedy} and Marginal Plots. This fact justifies why the ALE plots have the Moderate Severity in Table \ref{methods}. 
    
    \vsp
    \item \textbf{Use a Set of Non-Processed Features.} For the particular case of Feature Dependencies, the relationship that exists between features is deterministic and acknowledged. This time it is not a mitigation strategy but a full solution for preventing inconsistent artificial observations that appear as a result of dependencies. The solution consists of only modifying non-processed features and then do the preprocessing. This way, all the corresponding dependent features are changed correctly. For example, if the perturbation is done before binning or one-hot encoding, artificial observations remain consistent. Whereas, for the case of expert values, they get also modified if the dependent feature was perturbed. 
\end{enumerate}

\subsection{Contrastive Explanations}
\label{contrastive_sub}
Since humans tend to make comparisons between observations' predictions, Lipton \cite{lipton1990contrastive} states that explanations need to be contrastive by comparing a focal instance against a small reference group. Sometimes, it is desired to find a cluster of close by instances that share similar predictions. Other times, they are contrasted against similar instances but with different prediction outputs---these are counterfactual explanations. They seek slight changes---or perturbations---that the input should have had to obtain a different label. In the household's context, a refused candidate would like to know how its application could have been different in order to have been selected. In this sense, the feature's effect is not interpreted as contributions. A discussion of the challenge of Unreliable Counterfactuals and a corresponding mitigation strategy can be found at \cite{laugel2019dangers}. This strategy justifies it for having Moderate Severity for challenge \ref{unlikely} in Table \ref{methods}.

\vsp
\textbf{Mitigation Strategies.}
\begin{enumerate}
        
    \vsp    
    \item \textbf{Use of Filters by Class}. For Perturbation Approach methods, contrastivity is not straightforward. A strategy for mitigating this challenge is to adjust the procedure in which the artificial instances are generated. The strategy consists of only using values that appear, not only in the distribution data set, but also in the subset to which it is desired to contrast the focal observation. For example, in the household context, to contrast a focal household against those who were elected for the program, the extremely poor class. In this sense, the feature's effect is computed by using artificial instances values replacements derived from this particular class, the contrastive class of interest. An implementation of this strategy is discussed in Section \ref{add_contrastive}.
\end{enumerate}

\subsection{Non-Interpretable Model Input} 
\label{non}
Despite instances having an interpretative representation initially, their preprocessed representation, which the machine receives as input, is likely to be non-interpretable. For example, expert variables such as logarithms or entropies are non-interpretable features because, although they improve the model's accuracy, they are not appropriate for explaining the model to people. Also, binary features that were produced after a binning or a One-Hot Encoding process are useless for interpretation. 

\vsp
\textbf{Mitigation Strategies.} 
\begin{enumerate}[label=\arabic*)]
    \vsp
    \item \textbf{Use Aggregation of Effects.} This strategy only applies to methods that follow a Perturbation approach, and when procedures such as binning, grouping, or One-Hot Encoding are used. The available methodologies will compute effects feature-wise, particularly for each set of non-interpretable features derived from corresponding initial features. For the latter, it is assumed that these are interpretable. The mitigation strategy consists of aggregating non-interpretable effects, providing a single value corresponding to the interpretable feature's overall effect.
    
    \vsp
    \item \textbf{Use an Interpretable Representation.} A similar mitigation to the one discussed for dependencies can be for contrastiveness. It consists of selecting ad-hoc, a set of interpretable features, and transforming instances into an \textit{interpretable representation}. Alternatively, it consists of finding a transformation of the Feature Space into an Interpretable Space. The contrastivity would occur between nearby instances in the Interpretable Representation Feature space. Because this strategy could be straightforwardly implemented in Contrastive Approach Methods, Moderate Severity is given to this class for Challenge \ref{non} in Table \ref{methods}.
\end{enumerate}

\section{Proposed Methodology}
\label{proposed}

Sections \ref{uni_sub} to Section \ref{importances_sub} described in detail a model-agnostic methodology we developed for providing individual explanations in this application context. It follows a perturbation approach similar to some extent to the methods exposed in Table \ref{methods}, but it also gives a contrastive interpretability. Section \ref{uni_sub} describes the baseline methodology, its explanation outputs, and pseudocode for a straightforward implementation. The following three sections, from Section \ref{add_conditional} to Section \ref{add_contrastive}, each correspond to a challenge and mitigation strategy mentioned in \ref{unlikely} to \ref{non}. Throughout these sections, the baseline algorithm keeps enhancing, turning into a more robust method. Finally, Section \ref{importances_sub} describes the methodology for improving easy interpretability by adding a contrastive interpretation between the focal instance and a reference group (e.g., households in poverty).

\subsection{Features' Importances by Univariate Perturbation}
\label{uni_sub}

In this section, the beta-algorithm---the simplest version of the methodology discussed through Sections \ref{uni_sub} to \ref{add_contrastive}---is outlined. It follows a perturbation approach so the explanations are feature-wise effects.

\vsp
\textbf{Inputs.}
\begin{itemize}
    \item The complete household database $X$. It contains $n$ households (rows) and $d$ features (columns).
    \item The model $M$ that takes a household as input and outputs the Income Per Capita (IPC).
    \item A focal household $h$---the household of interest---represented as a vector with $d$ features.  
\end{itemize}

\textbf{Output.} The output of Algorithm \ref{uni_alg} is a vector where the $j$th element corresponds to the effect over the IPC of the $j$th feature.

\vsp
The Algorithm \ref{uni_alg} steps are described as follows. After predicting the IPC $\hat{y}$ of the focal household $h$, an iterative process for calculating each of the $d$ features' effects is initiated. The set $V_j$ of unique values that appear in the $j$th column of $X$ is defined. Then, each value $v \in V_j$, is used to perturb once the focal instance $h$. In each iteration, the artificial instance univariately generated is denoted by $h'$. Then, its IPC is predicted and denoted by $\hat{y}'$. Note that the values that we use for modifying the $j$th feature correspond to those that appear in the observation data set. This corresponds to the Mitigation Strategy 1) from Section \ref{unlikely}. The difference $\hat{y} - \hat{y}'$, denoted by $\Delta_j(v)$, can be interpreted as the monetary contribution of having that particular feature value different. Finally, to obtain a global measure of this effect, the weighted average among the different possible $\Delta_j(v)$ values is taken multiplied by the weight $w_j(v)$, which is the proportion in $X[\cdot, j]$ that $v$ appears. In this way, the effects of $\Delta_j(v)$ have more weight if $v$ has more representation in the database. Note that there is a value $v$ that will coincide with the value of $h$ and therefore, $\Delta_j(v) = 0$. By taking this weights average, the effects are being contrastive towards the whole database.

\vsp
\begin{algorithm}[ht!]
\caption{Features' Importances by \textit{univariate} perturbation}
\label{uni_alg}
\KwIn{
\begin{itemize}
    \item Data set $X \in \mathbb{R}^{n \times d}$
    \item Income Predictor Model $M: \mathbb{R}^{1 \times d} \rightarrow \mathbb{R}$
    \item Focal household $h \in \mathbb{R}^{1 \times d}$
\end{itemize}
}
\KwOut{Importance vector $I_{h}:  \mathbb{R}^{1 \times d} \rightarrow \mathbb{R}^{1 \times d}$ }

$\hat{y} \leftarrow M(h)$ \hspace{9.2cm} \textit{\# predict income}\\
\For{$j = 1,2, \dots, d$}{
    $V_{j} \leftarrow$ $\{X[\cdot, j]\}$ \hspace{7.8cm} \textit{\# set of unique values}\\
    \For{$v \in V_j$}{
        $h' \leftarrow h; \ \ h'[j] \leftarrow v$ \hspace{6.4cm} \textit{\# perturb one feature of} $h$\\
        $\hat{y}' \leftarrow M(h')$ \hspace{7.65cm} \textit{\# predict income}\\
        $\Delta_j(v) \leftarrow \hat{y} - \hat{y}'$  \hspace{6.96cm} \textit{\# difference between predictions}\\
        $w_j(v) \leftarrow $ \texttt{proportion of} $v$ \texttt{in} $X[\cdot, j]$
    }
    $ I_{h}(j) \leftarrow \sum_{v \in V_j}w_j(v) \Delta_j(v)$ \hspace{5.60cm} \textit{\# weighted average}\\
  }
\end{algorithm}

\subsection{Adding Ad-Hoc Conditionalities}
\label{add_conditional}
Unfortunately, the univariate approach from Algorithm \ref{uni_alg} generated impossible or extremely unlikely observations. To mitigate this problem the strategy 1)(i) from Section \ref{unlikely} was implemented. This strategy tries to avoid the generation of impossible observations by adding ad-hoc conditionalities which helps identifying incompatible feature values after the univariate perturbation. The ad-hoc conditionalities are defined as features subsets which we know ad-hoc are highly correlated. After perturbing univariately a feature value, the other features' from the shared subset undergo a procedure which consists of replacing their value with other values that have positive probability in the conditional distribution given by observing the univariate replacement value.

\begin{algorithm}[ht!]
\caption{Features' importances by \textit{conditional}, and \textit{univariate} perturbation}
\label{conditional_alg}
\KwIn{
\begin{itemize}
    \item Data sets $X \in \mathbb{R}^{n \times d}$, $Y \in \mathbb{R}^{n \times 1}$
    \item Income Predictor Model $M: \mathbb{R}^{1 \times d} \rightarrow \mathbb{R}$
    \item Focal household $h \in \mathbb{R}^{1 \times d}$
    \item Ad-hoc conditionalities $\subseteq \mathbb{R}^{1 \times d}$
\end{itemize}
}
\KwOut{Importance vector $I_{h}:  \mathbb{R}^{1 \times d} \rightarrow \mathbb{R}^{1 \times d}$ }
$\hat{y} \leftarrow M(h)$\hspace{9.85cm} \textit{\# predict income}\\
\For{$j = 1,2, \dots, d$}{
    $V_{j} \leftarrow$ $\{X[\cdot, j]\}$ \hspace{8.3cm} \textit{\# set of unique values}\\
    \For{$v \in V_j$}{
        $h' \leftarrow h; \ \ h'[j] \leftarrow v$ \hspace{6.9cm} \textit{\# perturb one feature of} $h$\\
        $h' \leftarrow \texttt{perturbation by ad-hoc cond. of }h'[j]=v$\\
        $\hat{y}' \leftarrow M(h')$ \hspace{8.2cm} \textit{\# predict income}\\
        $\Delta_j(v) \leftarrow \hat{y} - \hat{y}'$  \hspace{7.6cm} \textit{\# difference between predictions}\\
        $w_j(v) \leftarrow $ \texttt{proportion of} $v$ \texttt{in} $X[\cdot, j]$
    }
    $ I_{h}(j) \leftarrow \sum_{v \in V_j}w_j(v) \Delta_j(v)$ \hspace{6.15cm} \textit{\# weighted average}\\
}
\end{algorithm}

\vsp
\textbf{Mitigation Examples}
In our use case, perturbing univariately features, such as a household member's schooling, could yield impossible observations, such as ten-year-old children with graduate degrees.  To mitigate this challenge, the population was partitioned into eight different subsets given by four distinct age groups and by the household member's formal-sector economic activity. After the univariate perturbation was performed, the particular subset to which that observation belonged was identified. Then the conditional probabilities were used to perturb other highly correlated features. 

\vsp
After the implementation of Algorithm \ref{conditional_alg}, the problem for generating impossible observations was drastically mitigated. For example, if the age of a person with a graduate degree is perturbed to a ten-year-old, then all other features that could be incompatible with being a child were also modified.

\subsection{Adding Bivariate Perturbations}
\label{bi_sub}

Bivariate interactions across features were also considered, and thus the Mitigation Strategy 1)(ii) described in Section \ref{unlikely} was implemented. This approach weights the artificial instances accordingly to the probability of observing the two replaced values $v_j$ and $v_k$. Thus, artificial instances that are more likely to be observed weight more. The only differences between Algorithm \ref{bi_alg} and Algorithm \ref{conditional_alg} is that all possible feature pairs are used for generating artificial instances; and that the weighted average is not univariate but bivariate. Note that this bivariate is an extension of the univariate case, because when $j = k$, then $v_j = v_k$ perturbations are done univariately.

\begin{algorithm}[ht!]
\caption{Features' importances by \textit{conditional}, and \textit{bivariate} perturbation}
\label{bi_alg}
\KwIn{
\begin{itemize}
    \item Data sets $X \in \mathbb{R}^{n \times d}$, $Y \in \mathbb{R}^{n \times 1}$
    \item Income Predictor Model $M: \mathbb{R}^{1 \times d} \rightarrow \mathbb{R}$
    \item Focal household $h \in \mathbb{R}^{1 \times d}$
    \item Ad-hoc conditionalities $\subseteq \mathbb{R}^{1 \times d}$
\end{itemize}
}
\KwOut{Importance vector $I_{h}:  \mathbb{R}^{1 \times d} \rightarrow \mathbb{R}^{1 \times d}$ }

$\hat{y} \leftarrow M(h)$\hspace{9.8cm} \textit{\# predict income}\\
\For{$j = 1,2, \dots, d$}{
     $V_{j} \leftarrow$ $\{X[\cdot, j]\}$ \hspace{8.2cm} \textit{\# set of unique values}
}

\For{$(j,k) = 1,2, \dots, d$}{
    \For{each : $v_j \in V_j$ and $v_k \in V_k$}{
        $h' \leftarrow h; \ \ h'[j] \leftarrow v_j; \ \  h'[k] \leftarrow v_k$ \hspace{4.4cm} \textit{\# perturb two features of} $h$\\
        $h' \leftarrow \texttt{perturbation by ad-hoc cond. of }h'[j]=v_j \texttt{ and } h'[k]=v_k$\\
        $\hat{y}' \leftarrow M(h')$ \hspace{8.15cm} \textit{\# predict income}\\
        $\Delta_{j,k}(v_j, v_k) \leftarrow \hat{y} - \hat{y}'$  \hspace{6.5cm} \textit{\# difference between predictions}\\
        $w_{j,k}(v_j, v_k) \leftarrow $ \texttt{proportion of} $(v_j, v_k)$ \texttt{in} $X[\cdot, j]$
    }
    $ I_{h}(j) \leftarrow \sum_{v_j \in V_j} \frac{1}{d}(\sum_{v_k \in V_k} w_{j,k}(v) \Delta_{j,k}(v))$ \hspace{3.55cm} \textit{\# weighted average}\\
}
\end{algorithm}

\vsp
\textbf{Mitigation Examples}
The improvement of the implementation of Algorithm \ref{bi_alg} is not as straightforward to observe as in Algorithm \ref{conditional_alg}. But with this approach, the artificial observations generated by an unlikely bivariate replacement, assuming this combination was not invalidated by ad-hoc conditionalities, would still be given a low weight. For example, the number of household members is correlated with the number of bedrooms. If the artificial instance has ten members, we would expect a high number of bedrooms. This strategy weights proportionally the instances given the number of bedrooms and household members, for the case when these features correspond to $v_j$ and $v_k$. Another example is the household's area with the number of rooms or the years of education with the highest educational attainment level.

\subsection{Adding a Contrastive Interpretation}
\label{add_contrastive}
As mentioned in Section \ref{real_sub}, the Perturbation Approach methodologies are not inherently contrastive. One of the challenges enlisted is that it is a desirable property to make this explanations comparable with a group of reference. To enable the methodology so far described with a contrastive explanation interpretation we implemented strategy 1) discussed in Section \ref{contrastive_sub}. Algorithm \ref{contrastive_alg} is practically identical to Algorithm \ref{bi_alg}, except for a few altered lines. Note that in line 1 we now define $X_p$, the filtered database of households below the poverty line. To define the filter, we use the ground-truth income $Y$, and compare it against the constant \texttt{poverty line} value. Everything else remains the same, although we now use $X_p$ instead of $X$ throughout the rest of the algorithm.

\begin{algorithm}[ht!]
\caption{Features' Importances by \textit{conditional, bivariate}, and \textit{contrastive perturbation}}
\label{contrastive_alg}
\KwIn{
\begin{itemize}
    \item Data sets $X \in \mathbb{R}^{n \times d}$, $Y \in \mathbb{R}^{n \times 1}$
    \item Income Predictor Model $M: \mathbb{R}^{1 \times d} \rightarrow \mathbb{R}$
    \item Focal household $h \in \mathbb{R}^{1 \times d}$
    \item Ad-hoc conditionalities $\subseteq \mathbb{R}^{1 \times d}$
\end{itemize}
}
\KwOut{Importance vector $I_{h}:  \mathbb{R}^{1 \times d} \rightarrow \mathbb{R}^{1 \times d}$ }
$X_p \leftarrow \texttt{filter households } X[i, \cdot] \texttt{ such that } Y_i < \texttt{poverty line} $ \\
$\hat{y} \leftarrow M(h)$\hspace{9.9cm} \textit{\# predict income}\\
\For{$j = 1,2, \dots, d$}{
     $V_{j} \leftarrow$ $\{X_p[\cdot, j]\}$ \hspace{8.2cm} \textit{\# set of unique values}
}

\For{$(j,k) = 1,2, \dots, d$}{
    \For{each : $v_j \in V_j$ and $v_k \in V_k$}{
        $h' \leftarrow h; \ \ h'[j] \leftarrow v_j; \ \  h'[k] \leftarrow v_k$ \hspace{4.5cm} \textit{\# perturb two features of} $h$\\
        $h' \leftarrow \texttt{perturbation by ad-hoc cond. of }h'[j]=v_j \texttt{ and } h'[k]=v_k$\\
        $\hat{y}' \leftarrow M(h')$ \hspace{8.2cm} \textit{\# predict income}\\
        $\Delta_{j,k}(v_j, v_k) \leftarrow \hat{y} - \hat{y}'$  \hspace{6.6cm} \textit{\# difference between predictions}\\
        $w_{j,k}(v_j, v_k) \leftarrow $ \texttt{proportion of} $(v_j, v_k)$ \texttt{in} $X_p[\cdot, j]$
    }
    $ I_{h}(j) \leftarrow \sum_{v_j \in V_j} \frac{1}{d}(\sum_{v_k \in V_k} w_{j,k}(v) \Delta_{j,k}(v))$ \hspace{3.6cm} \textit{\# weighted average}\\
}
\end{algorithm}

\vsp
\textbf{Mitigation Examples}

In the case study it was desirable to provide contrastive interpretations. We wished to contrast any focal house with a reference group, which in this case were those households that lived below the poverty line. After the implementation, instead of interpreting the importance as the monetary contribution of having that particular feature value different to the average household in the entire population, it is seen as the contribution against a household that lives in poverty. This comparison is suitable for understanding the model's estimation when a focal household contains feature-values that do not accord with those typically seen in the reference group.

\subsection{Importances of Feature Groups}
\label{importances_sub}

In this section, an algorithm for obtaining importances at the \textit{feature groups} level, instead of at the \textit{feature} level, is described. A \textit{feature group} is a collection of common features such as ``assets'', ``sociodemographic characteristics'', ``occupation characteristics'', or ``building/household characteristics''. The feature group's importance of a given household is defined as the average of feature importances for features in the feature group. These feature groups' importances are useful for contrasting group-wise new focal households with the group of reference---the extremely poor---in a simple and summarized manner. Once we obtain all the feature groups' importances of the contrastive set and of the focal household, percentiles can be used to show how the focal instance compares to the contrastive set in terms of each of the feature groups (e.g., household assets).

\vsp
Algorithm \ref{groups_alg} describes how to obtain the feature groups' importances of all the households in the contrastive set.\footnote{How to implement the use of percentiles for contrasting a new focal is not described here, but it is straightforward.} Algorithm \ref{groups_alg} does not assume that feature importances were obtained previously, so it explicitly gets them via Algorithm \ref{contrastive_alg}.

\begin{algorithm}[ht!]
\caption{Groups' Importances by \textit{conditional, bivariate}, and \textit{contrastive perturbation}}
\label{groups_alg}

\KwIn{
\begin{itemize}
    \item Data sets $X \in \mathbb{R}^{n \times d}$, $Y \in \mathbb{R}^{n \times 1}$
    \item Income Predictor Model $M: \mathbb{R}^{1 \times d} \rightarrow \mathbb{R}$
    \item Ad-hoc conditionalities $C_j$ for each $j$th feature, $j = 1,2 \dots d$
    \item Indices sets $K_k$ where $j \in K_k$ iff the $j$th feature is of the $k$th \textit{group}.
\end{itemize}
}
\KwOut{Importance vector $I_{h}:  \mathbb{R}^{1 \times d} \rightarrow \mathbb{R}^{1 \times d}$ }
$X_p \leftarrow \texttt{filter households } X[i, \cdot] \texttt{ such that } Y_i < \texttt{poverty line} $ \\
$m \leftarrow \vert X_p \vert$  \hspace{7.2cm}  \textit{\# cardinality of contrastive set}\\
\For{$k = 1,2, \dots, p$}{
    $K_k \leftarrow \{j: $ the $j$th feature is of the $k$th group$\}$ \hspace{0.2cm} \textit{\# define index set}\\
    $q_k \leftarrow |K_k|$ \hspace{6.5cm}          \textit{\# save cardinality}
}
\For{$i = 1,2, \dots, m$}{
    $h_i \leftarrow X_p[i, \cdot]$  \hspace{6.1cm}  \textit{\# contrastive focal household of $i$th iteration}\\
    $I_{h_i} \leftarrow$ Algorithm \ref{contrastive_alg}$(X, Y, M, h_i, \{C_j \}_j)$  \hspace{1.5cm} \textit{\# feature importances  vector for focal household} \\
    \For{$k= 1,2, \dots , p$}{
        $\mathcal{I}_{h_i}(k) \leftarrow \frac{1}{q_k}  \sum_{j \in K_k} I_{h_i}(j)$ \hspace{3.1cm} \textit{\# group importances vector for focal household}
    }
}
\end{algorithm}

\pagebreak
\section{A Visual and Interactive Explainability Platform}
\label{platform}

In our experience, any interpretability methodology needs a visual interface to achieve true usefulness for technical and non-technical decision-makers. Hence, for our application domain, we developed an interactive platform that provides a visual explanation of the statistical rationales underlying the poverty assessment of any given household. The platform analyzes individual cases applying the interpretable explanation methodologies described in previous sections of this document, and visualizes the main sociodemographic characteristics that determine the poverty classification of households according to the model. Moreover, the platform can be adapted to any estimation model.

\begin{figure}[ht!]
    \centering
    \includegraphics[scale=0.55]{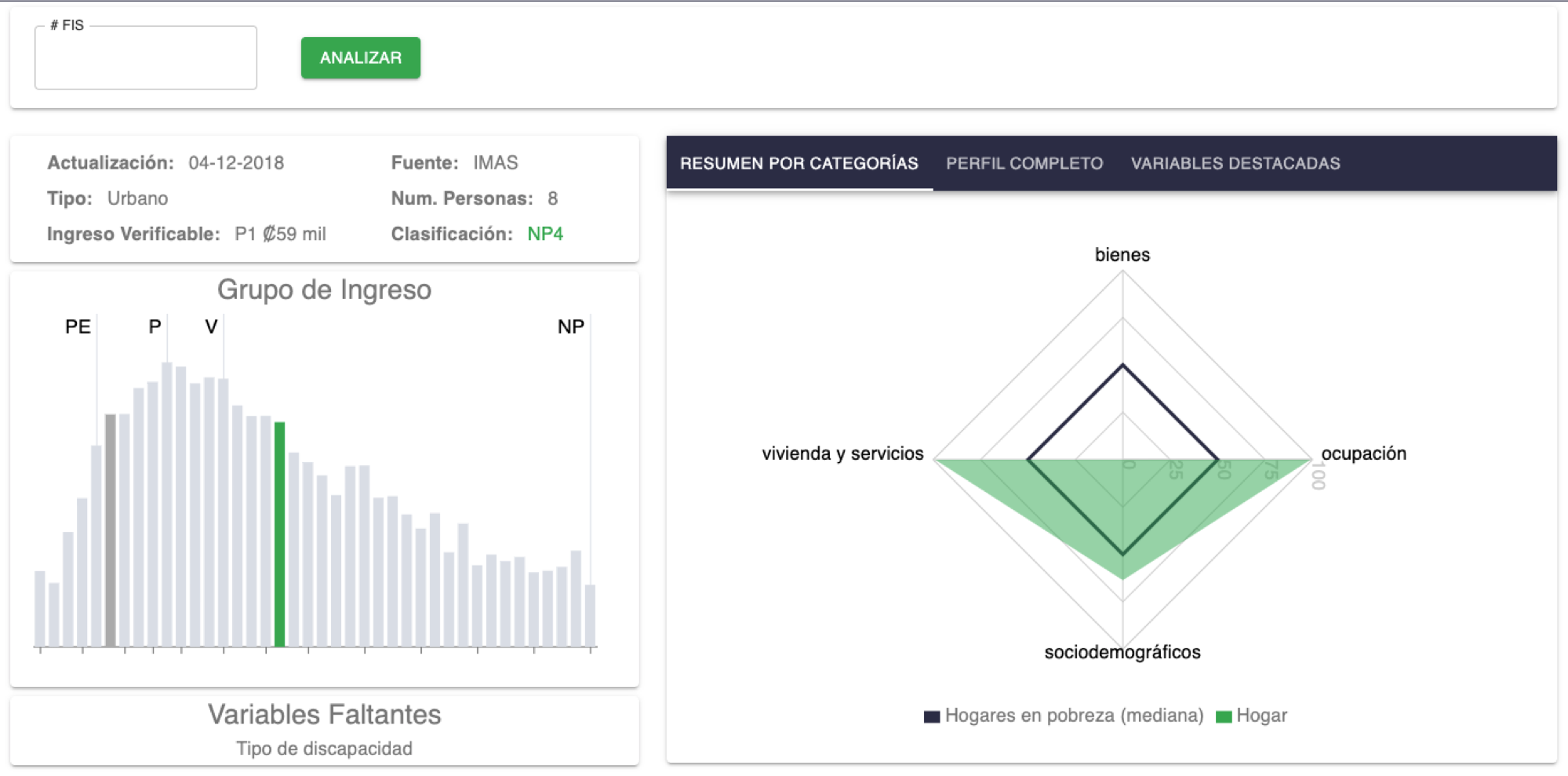}
    \caption{Visual interface of the platform.}
    \label{platform_img}
\end{figure}

\vsp
In Figure \ref{platform_img}, we show the the interface of the platform. The left panel shows the distribution of per-capita household income in the country, and highlights the estimated income level (including formal and informal income, green bar) as well as the formal income observed (grey bar) for the focal household. Moreover, the platform shows some key contextual variables relevant for understanding the analysis of the focal household, including the date when the information was collected (e.g., to identify issues with outdated information), as well as a list of variables with missing values (which can be associated with estimation issues).

\vsp
On the right panel, via a radar plot, the platform summarizes the sociodemographic profile of the focal household and how it relates to income poverty according to the estimation model (following the feature group importances methodology described in Section \ref{importances_sub}). In particular, the radar compares the values of the focal household against the median of households in poverty along each relevant category of variables: housing and services, assets, occupation, and sociodemographic. For example, the radar plot might show that a household is classified as non-poor because its assets denote higher income than the median household in poverty (or, for example, higher that 95\% of households in poverty). 

\vsp
The platform is currently being used by social institutions at the national-level in a Costa Rica, to assist social workers in the analysis and assessment of households' socio-economic levels. Moreover, the platform is often used to synergize the statistical rationales of the estimation model with additional qualitative information available to social workers.

\section{Conclusion}
\label{conclusion}

In this case study, we exposed some of the challenges that arise when using post-hoc methods in real-world scenarios, as well as a set of useful strategies to address them. The list of methods is not exhaustive but emphasizes the most prominent challenges faced when attempting to generate a layer of explainability in application domains like ours, and particularly regarding a broad and prominent class of methods: perturbation-based approaches. This work shows that unlikely or impossible observations appear as potentially the most relevant challenge, as it affects the majority of methods, can cause unreliable and counter-intuitive explanations, and most available mitigation strategies offer only partial solutions.

\vsp
In particular, we defined an ad-hoc methodology for providing an explanation layer to arbitrary ML models, in the context of poverty predictions and their use for prioritizing access to social policies. The methodology starts with a perturbation-based approach, and provides ad-hoc mitigation strategies to the most relevant challenges faced by this relevant type of approach.

\vsp
In this work we have also highlighted that \textit{contrastive} methods, such as counterfactual-based and \textit{prototype and criticism} methods, can naturally circumvent many challenges akin to the perturbation-based approaches, but potentially face others, such as the curse of dimensionality, and NP-hard optimization problems. Future work may examine the implementation of contrastive approaches in the context of poverty estimation, and report in detail their advantages and challenges.

\vsp
We hope this work is useful for practitioners seeking to implement explainability layers into complex statistical models that influence decisions in real-world contexts. We also wish the work helps stimulate applied work in academia for the development and evaluation of explainability methods that more naturally address common challenges faced by practitioners.

\pagebreak
\begin{multicols}{2}

\bibliographystyle{unsrt}
\bibliography{main}
\end{multicols}

\end{document}